\documentclass{article} 
\usepackage{booktabs}

\usepackage{amsmath,amsfonts,bm}

\def\eqref#1{equation~\ref{#1}}

\def\1{\bm{1}}

\DeclareMathAlphabet{\mathsfit}{\encodingdefault}{\sfdefault}{m}{sl}
\SetMathAlphabet{\mathsfit}{bold}{\encodingdefault}{\sfdefault}{bx}{n}

\usepackage{float}
\usepackage{hyperref}
\usepackage{url}
\usepackage{placeins}
\usepackage{graphicx}
\usepackage{iclr2026_conference,times}
\usepackage{amssymb}
\usepackage{mathtools}
\title{Does Semantic Noise Initialization Transfer from Images to Videos? A Paired Diagnostic Study}

\author{
Yixiao Jing$^{1,*}$ \quad
Chaoyu Zhang$^{1,*}$ \quad
Zixuan Zhong$^{2,*}$ \quad
Peizhou Huang$^{1,\dagger}$ \\
$^{1}$University of Michigan \qquad
$^{2}$University College London \\
\texttt{peizhou@umich.edu}
}

\iclrfinalcopy 
\begin{document}

\maketitle

\begin{abstract}
Semantic noise initialization has been reported to improve robustness and controllability in image diffusion models. Whether these gains transfer to text-to-video (T2V) generation remains unclear, since temporal coupling can introduce extra degrees of freedom and instability. We benchmark semantic noise initialization against standard Gaussian noise using a frozen VideoCrafter-style T2V diffusion backbone and VBench on 100 prompts. Using prompt-level paired tests with bootstrap confidence intervals and a sign-flip permutation test, we observe a small positive trend on temporal-related dimensions; however, the 95\% confidence interval includes zero ($p \approx 0.17$) and the overall score remains on par with the baseline. To understand this outcome, we analyze the induced perturbations in noise space and find patterns consistent with weak or unstable signal. We recommend prompt-level paired evaluation and noise space diagnostics as standard practice when studying initialization schemes for T2V diffusion. Our code and evaluation scripts are available at: \url{https://github.com/klkds/golden-noise-transfer}.
\end{abstract}
\section{Introduction}
\label{intro}

Text-to-video (T2V) diffusion models are sensitive to random seeds: different initial Gaussian noises can yield large semantic and motion variations under the same prompt, complicating controllability and reliable comparison\cite{singh2025demystifying}.
Recent image-generation work suggests that teacher-aligned noise initialization can improve robustness by moving the starting noise distribution closer to a teacher's preferred region in noise space\cite{ahn2024noise, li2025noisear, eyring2025noise}.
A natural hypothesis is that videos may benefit even more, since temporal dynamics amplify seed-induced variance\cite{ho2022imagen}.

In this work, we conduct a focused diagnostic study on transferring semantic noise initialization from image generation to video diffusion.
We instantiate a lightweight noise mapper (named NPNet) on top of a frozen video diffusion backbone (VideoCrafter-style) and evaluate on 100 prompts using VBench.
Crucially, we report the paired statistical testing over prompts (bootstrap CI and permutation test), which is necessary when effect sizes are small relative to prompt-level variance\cite{efron1992bootstrap, dror2018hitchhiker}.
We further analyze the underlying mechanisms that govern how semantic (“golden”) noise perturbations propagate in the spatio-temporal diffusion process, by conducting a cross-model diagnostic on Open-Sora2 and VideoCrafter, which employ different video diffusion sampling mechanisms.

\noindent \textbf{Key findings.} 
(i) On VideoCrafter, semantic noise mapping shows a slight but statistically insignificant positive trend on temporal metrics ($p \approx 0.17$), with overall scores remaining on par with the baseline. 
(ii) Prompt-level variance dominates the effect size, placing this approach in a low signal-to-noise ratio (low-SNR) regime. 
(iii) Noise-space diagnostics indicate that while induced semantic perturbations are structured, their directional stability and spatiotemporal frequency profiles vary substantially between Open-Sora2 and VideoCrafter, helping explain the inconsistent temporal gains across models.

\vspace{1em}
\noindent \textbf{Contributions.}
\begin{itemize}
  \item Reproducible paired evaluation of semantic noise initialization on a VideoCrafter-style T2V diffusion model over 100 prompts.
  \item Prompt-level significance testing (bootstrap CI and permutation test), clarifying that temporal-metric trends are not statistically reliable under this setting.
  \item We develop cross-model noise-space diagnostics that characterize the directional stability and spatiotemporal frequency structure of semantic perturbations, enabling systematic comparison across video diffusion backbones.
\end{itemize}

\section{Related Work}
\label{related_works}

\noindent \textbf{Diffusion-based T2V Generation.}
Diffusion-based generators are a dominant paradigm for high-fidelity image and video synthesis.
Modern T2V diffusion largely inherits DDPM and latent-diffusion formulations, injecting text conditioning via cross-attention and classifier-free guidance (CFG)~\cite{ho2020denoising,rombach2022high,ho2022classifier}.
Recent systems (e.g., VideoCrafter-style backbones) extend denoising to spatio-temporal latents to model motion and temporal coherence~\cite{chen2023videocrafter1}, but typically keep the highest-noise initialization as an isotropic Gaussian draw, leaving \emph{initialization} relatively underexplored.

\noindent \textbf{Seed Sensitivity and Noise-space Learning.}
T2V diffusion is sensitive to random seeds, motivating noise-space transformations on frozen backbones, exemplified by Golden Noise: semantically aligned noise targets are constructed with a teacher diffusion model and a lightweight mapper is trained from standard Gaussian noise to the learned distribution~\cite{zhou2025golden}.
Inversion methods (commonly DDIM inversion) provide practical trajectories between noise and denoised latents for target construction~\cite{song2020denoising}.
Our semantic noise initialization follows this teacher-in-noise principle and adapts it to spatio-temporal video latents to reduce seed-induced variance without retraining the full backbone.

\noindent \textbf{Evaluation of T2V Quality.}
T2V evaluation typically combines semantic alignment, which is often CLIP-based~\cite{radford2021learning}, perceptual similarity and diversity (e.g., Learned Perceptual Image Patch Similarity, LPIPS)~\cite{zhang2018unreasonable}, and other temporal consistency metrics; suites on benchmarks such as VBench offer standardized protocols~\cite{huang2024vbench}.
To isolate the impact of initialization, we keep the backbone, sampler, prompt, and seed fixed, and make changes only the initial noise so that differences in temporal behavior can be attributed to the learned initializer.

\section{Methodology}
\label{method}

\noindent \textbf{Problem Setup.}
Let $G_{\theta}$ denote a frozen T2V diffusion generator that samples a video from an initial noise latent $z_T \sim \mathcal{N}(0,I)$ and a prompt $p$.
Seed sensitivity arises because perturbations in $z_T$ can induce variations in the generated spatio-temporal latent trajectory.

\noindent \textbf{Semantic (Golden) Noise Targets.}
Following the teacher-in-noise principle, we consider a semantic target noise $z_T^{\star}$ (golden noise) that is more aligned with the prompt-conditioned denoising trajectory.
In practice, $z_T^{\star}$ can be obtained by optimization or inversion-style procedures that search in noise space for an initialization yielding higher semantic and temporal quality under a fixed backbone and sampler.
This extraction is computationally non-trivial for videos, as evaluating a candidate noise typically requires running the spatio-temporal denoising process.

\noindent \textbf{NPNet: A Lightweight Noise Mapper.}
We train a lightweight mapper $f_{\phi}$ that transforms standard Gaussian noise into a semantic initialization:

\begin{equation}
\hat{z}_T = f_{\phi}(z_T, p),
\label{eq:zhat0}
\end{equation}

where $f_{\phi}$ is conditioned on the prompt \(p\) (e.g., via text embedding injection).
We train $f_{\phi}$ to approximate the extracted targets using a regression loss:

\begin{equation}
\mathcal{L}(\phi) = \mathbb{E}\left[\left\|f_{\phi}(z_T,p) - z_T^{\star}\right\|_2^2\right],
\label{eq:zhat1}
\end{equation}

while keeping $G_{\theta}$ frozen.
At inference, we replace $z_T$ with $\hat{z}_T$ and sample with the same backbone, sampler, and guidance.

\noindent \textbf{Cost Model and Practical Overhead.}
Unlike images, extracting $z_T^{\star}$ for videos can incur substantial overhead because each candidate requires spatio-temporal denoising evaluation.
This motivates reporting the cost--benefit trade-off explicitly: even when metric gains exist, they must outweigh the additional compute for target extraction and training.

\section{Experiments}
\label{exp}
\subsection{Setup}
Unless otherwise stated, we evaluate on 100 prompts sampled from the VBench prompt set with 5 random seeds per prompt.
For each prompt--seed pair, we keep the text prompt, the backbone $G_{\theta}$, the scheduler, and the CFG configuration identical, and change only the initial latent at the highest-noise timestep ($t=T$).
The baseline samples $z_T \sim \mathcal{N}(0,I)$, while NPNet deterministically maps the same sampled noise using the same prompt embedding,
$\hat{z}_T = f_{\phi}(z_T, p)$, and then runs the same sampling procedure thereafter.
Thus, both methods share the same conditioning in the diffusion backbone; the only difference is the initialization at $t=T$ (i.e., the initial latent fed into the sampler).
For statistical analysis, we use prompt-level paired comparisons: we first average each metric over the 5 seeds for each prompt, and then compute paired differences across the 100 prompts (i.e., the statistical unit is the prompt, $N=100$), unless explicitly noted otherwise.

\subsection{Quantitative evaluation}
We adopt VBench and use temporal\_style as the primary temporal metric, as it most directly captures temporal dynamics like flicker and jitter, and exhibits lower prompt-level noise than composite temporal aggregates in our setting.
To assess reliability across prompts, we perform prompt-level paired tests over the 100 prompts.
Concretely, for each prompt we average the metric over the 5 seeds for the baseline and for NPNet, then form a paired difference (NPNet minus baseline).
We report (i) a bootstrap confidence interval (CI) for the mean paired difference and (ii) a paired sign-flip permutation test for the null hypothesis of zero mean difference.

\begin{table}[h]
\centering
\caption{Mean VBench scores on 100 prompts.}
\label{tab:allmetrics}
\begin{tabular}{lccc}
\toprule
Dimension & Baseline & NPNet & $\Delta$ \\
\midrule
aesthetic\_quality & 0.638083 & 0.634992 & -0.003091 \\
imaging\_quality & 0.715084 & 0.707932 & -0.007151 \\
background\_consistency & 0.976592 & 0.976812 & +0.000220 \\
subject\_consistency & 0.977814 & 0.978034 & +0.000220 \\
temporal\_style & 0.076961 & 0.078716 & +0.001754 \\
\bottomrule
\end{tabular}
\end{table}

Table~\ref{tab:allmetrics} reports seed-averaged mean scores over prompts for reference.
All statistical claims are based on the prompt-level paired testing with prompt as the unit ($N=100$; shown in Appendix~\ref{app:sig}, Table~\ref{tab:sig_paired}); the improvement on temporal\_style is not significant (95\% CI crosses zero; $p=0.1687$). Although aggregate metrics are neutral and improvements are not statistically significant, we occasionally observe prompt-specific gains in fine textures (e.g., fur/scales); see Appendix~\ref{app:visuals}.

\subsection{Qualitative noise-space diagnostics}
To interpret the quantitative trade-off, we analyze the geometry and spatiotemporal frequency characteristics of golden noise $z_g$ relative to standard Gaussian noise $z$. Beyond characterizing VideoCrafter alone, we include a cross-model diagnostic
with Open-Sora2 to assess whether the induced noise structure is intrinsic or
dependent on the sampling dynamics.
We define the displacement $d=z_g-z$ and aggregate statistics over 100 prompts $\times$ 5 seeds. Formal definitions of the geometry- and frequency-based metrics are provided in Appendix~\ref{app:qual_metrics}.
\FloatBarrier
\begin{table}[h]
\centering
\caption{Global geometry and directional consistency of golden-induced perturbations for Open-Sora2 and VideoCrafter.}
\label{tab:geometry_direction_cross}
\begin{tabular}{lcc}
\toprule
Metric & Open-Sora2 & VideoCrafter \\
\midrule
$\lVert z_g - z \rVert / \lVert z \rVert$
& 0.022 & 0.110 \\
$\cos(z, z_g)$
& 0.9997 & 0.9939 \\
Directional Stability (DirStab)
& 0.631 & 0.200 \\
CV$_{\lVert d \rVert}$
& 0.064 & 0.110 \\
Explained Variance Ratio (EVR1)
& 0.464 & 0.343 \\
\bottomrule
\end{tabular}
\end{table}

\begin{table}[h]
\centering
\caption{Frequency summary: global invariance between $z$ and $z_g$, and
spatiotemporal characteristics of the displacement $d=z_g-z$.}
\label{tab:freq_combined_cross}
\begin{tabular}{lccc|ccc}
\toprule
& \multicolumn{3}{c}{Open-Sora2} 
& \multicolumn{3}{c}{VideoCrafter} \\
\cmidrule(lr){2-4} \cmidrule(lr){5-7}
Metric 
& Mean & P10 & P90
& Mean & P10 & P90 \\
\midrule
$\Delta$ Spatial HF ($z_g - z$)
& -0.00049 & -0.00114 & 0.00006
& -0.01489 & -0.02086 & -0.00745 \\
\midrule
$\mathrm{sp\_hf}(d)$
& 0.17996 & 0.06667 & 0.27212
& 0.24804 & 0.14243 & 0.42861 \\
$\mathrm{t\_hf}(d)$
& 0.85502 & 0.82826 & 0.87007
& 0.59898 & 0.45872 & 0.73720 \\
$\mathrm{tDiffRel}(d)$
& 0.60583 & 0.41060 & 0.75598
& 0.54638 & 0.45229 & 0.64628 \\
\bottomrule
\end{tabular}
\end{table}

\FloatBarrier

Table~\ref{tab:geometry_direction_cross} shows that $z_g$ stays close to $z$ in both models, but the
displacement $d=z_g-z$ is considerably more aligned across seeds in Open-Sora2 than in VideoCrafter (DirStab/EVR1).
Table~\ref{tab:freq_combined_cross} indicates near-invariant input frequency in Open-Sora2
($\Delta$ Spatial HF $\approx 0$) but a systematic frequency shift in VideoCrafter, while the displacement exhibits model-dependent spatiotemporal structure.

\section{Discussion}
Our paired evaluation indicates that semantic noise initialization yields at most a small positive trend on temporal-related VBench dimensions, but the effect is not statistically significant under 100 prompts.
The qualitative diagnostics help interpret this low-SNR outcome: in Open-Sora2, golden noise remains extremely close to the Gaussian prior in global geometry, yet induces a structured, prompt-conditioned displacement that is consistent across seeds (DirStab/EVR1).
Crucially, the change concentrates in the displacement $d=z_g-z$ rather than the global spectrum of $z_g$, and exhibits a spatiotemporal imbalance: spatially smooth but temporally high-frequency.

In contrast, when evaluated on VideoCrafter with DDIM sampling, the induced displacement becomes substantially more dispersed in direction across seeds. As a consequence, the temporal high-frequency components are less concentrated and their relative dominance is reduced, resulting in a weaker amplification of temporal instability. We attribute this effect to the path-dependent dynamics of DDIM, which tend to rotate and diffuse initial directional perturbations across sampling steps.

This offers a coherent mechanism for the observed trade-off: a stable low-frequency bias may support coarse temporal coherence, while temporally jittery components can amplify flicker/jitter through temporal coupling during denoising, degrading perceptual quality.
Taken together, these findings suggest that directly transferring image-style semantic/golden initialization to videos can enter a regime where the signal exists and is structured, but its temporal frequency characteristics make the net gain fragile under standard benchmark protocols.

\section{Conclusion}
We studied whether semantic (teacher-aligned) noise initialization transfers from images to T2V diffusion using a frozen video generation model's backbone and a lightweight prompt-conditioned mapper (NPNet).
On 100 prompts with prompt-level paired tests, NPNet shows a small positive trend on temporal-related metrics but no statistically significant improvement, leaving overall quality near parity with the Gaussian baseline.
Our noise-space diagnostics suggest the method mainly adds a structured spatiotemporal displacement rather than changing the global spectrum, which can yield fragile temporal gains and occasional perceptual degradation.

\newpage

\bibliographystyle{iclr2026_conference}
\bibliography{iclr2026_conference}


\appendix
\section{Additional Qualitative Results}
\label{app:visuals}
\begin{figure*}[h]
    \centering
    \includegraphics[width=\textwidth]{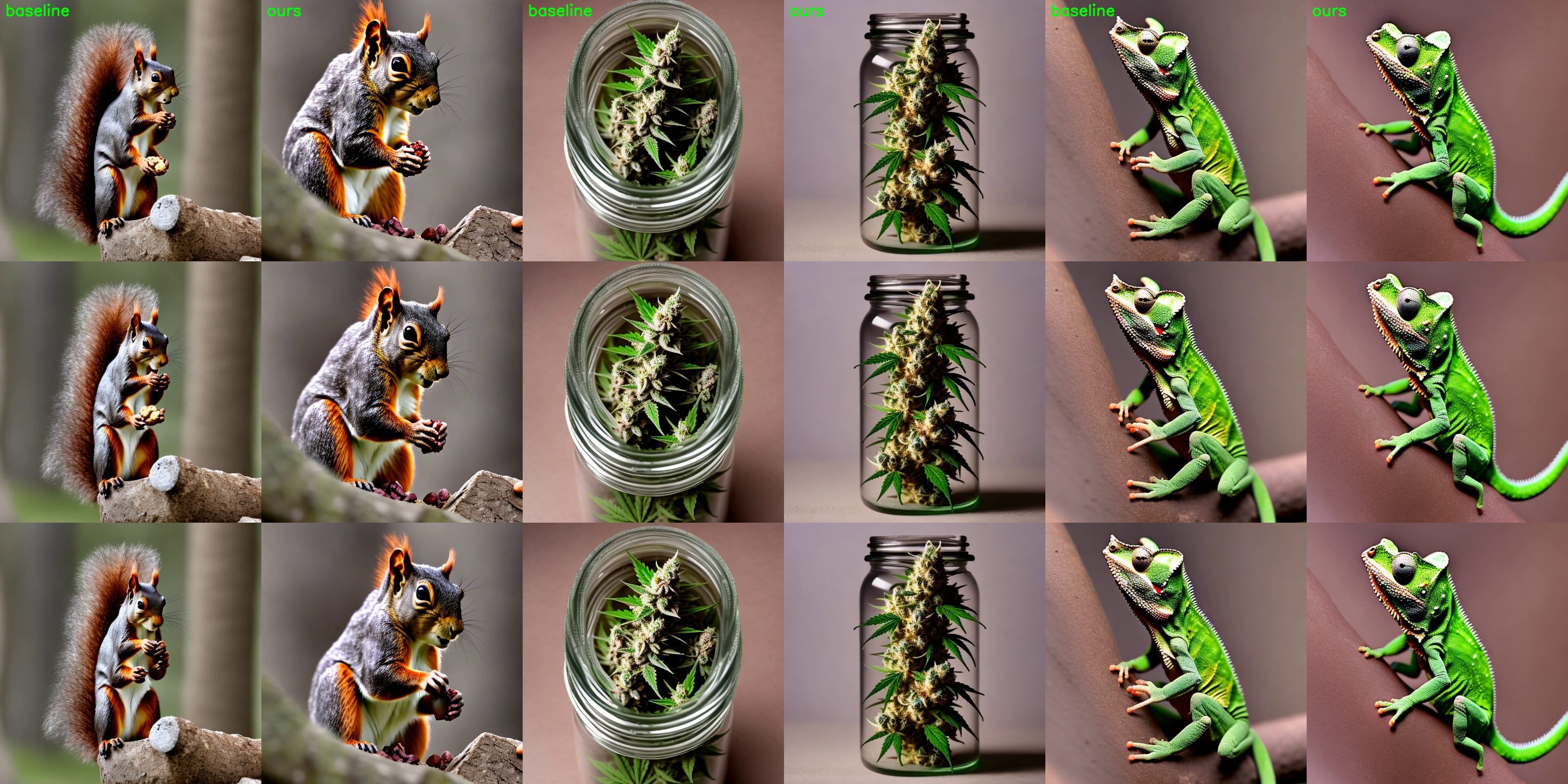} 
    \caption{\textbf{Qualitative comparison on VideoCrafter.} (Full size view) We visualize samples from Baseline (columns 1, 3, 5) versus our NPNet initialization (columns 2, 4, 6). Our method improves visual fidelity and detail consistency (e.g., the fur of the squirrel and the scales of the lizard) without changing the diffusion backbone.}
    \label{fig:qualitative_appendix}
\end{figure*}

In this section, we provide enlarged visual comparisons to supplement the quantitative results discussed in Section~\ref{exp}. Figure~\ref{fig:qualitative_appendix} contrasts the standard Gaussian initialization (Baseline) with our proposed semantic noise initialization (NPNet) across three distinct prompts. As observed, our method yields sharper high-frequency details and more consistent textures, particularly in challenging regions such as animal fur and skin scales.
\section{Limitations}
First, our conclusions are scoped to a VideoCrafter-style backbone and a fixed sampling/guidance configuration; different backbones, samplers, or guidance scales may shift the balance between temporal coherence and perceptual quality.
Second, our evaluation relies on VBench metrics and prompt-level paired testing; while this is appropriate for small effects, it may not fully capture human preference or prompt-specific failure modes (e.g., rare motion artifacts).
Third, the qualitative analysis focuses on aggregate noise-space statistics and frequency summaries; these diagnostics explain correlations but do not constitute a causal proof of how specific frequency components propagate through the denoising dynamics.
Finally, extracting or optimizing golden targets for video remains non-trivial in compute; thus, even when improvements exist, the overall cost--benefit trade-off may be unfavorable in practical deployment.

\section{Ethics Statement}
This work is a technical diagnostic study on existing open-source T2V diffusion models. It does not involve human subjects, private data collection, or crowdsourced annotation. We acknowledge the potential societal impact of generative video models (e.g., misinformation or bias), which stems from the pre-trained backbones used in our analysis rather than the proposed initialization method itself.

\section{Reproducibility Statement}
To ensure reproducibility, we build our experiments upon the open-source VideoCrafter backbone and the VBench evaluation suite. We have detailed the experimental setup, including the number of prompts, random seeds, and evaluation protocols, in Section 4 and the Appendix. The complete code, configuration files, and the list of prompts used in this study will be made publicly available on GitHub upon acceptance.
\section{Paired significance tests}
\label{app:sig}

To rigorously assess the reliability of the observed improvements, we perform a paired statistical analysis with the text prompt as the independent unit ($N=100$). Table~\ref{tab:sig_paired} details the results of both bootstrap confidence interval estimation (10,000 resamples) and a paired sign-flip permutation test. The analysis confirms that while the mean difference is positive, the improvement is not statistically significant at the $\alpha=0.05$ level ($p \approx 0.17$), indicating that the effect size is small relative to the inter-prompt variance.

\begin{table}[t]
\centering
\caption{Prompt-level paired significance analysis over 100 prompts (seed-averaged). Mean $\Delta$ is NPNet minus Baseline.}
\label{tab:sig_paired}
\begin{tabular}{lccccc}
\toprule
Metric & Baseline & NPNet & Mean $\Delta$ & 95\% CI (bootstrap) & $p$ (perm.) \\
\midrule
temporal\_style & 0.076961 & 0.078716 & +0.001754 & [$-0.000658$, $0.004166$] & 0.1687 \\
\bottomrule
\end{tabular}
\end{table}

\section{Additional VBench dimension scores}
\label{app:vbench}

Table~\ref{tab:allmetrics} reports global mean VBench scores over 100 prompts for completeness.
Statistical claims in the main text rely on prompt-level paired testing (Appendix~\ref{app:sig}, Table~\ref{tab:sig_paired}).

\section{Definitions of qualitative noise-space metrics}
\label{app:qual_metrics}

\noindent \textbf{Notation and scope.}
We analyze standard Gaussian initialization $z$ (\texttt{z\_T}) and the corresponding golden initialization $z_g$ (\texttt{z\_T\_target}) at the same diffusion timestep.
We define the displacement
\begin{equation}
d \triangleq z_g - z.
\label{eq:def_d_app}
\end{equation}
Unless stated otherwise, statistics are aggregated over 100 prompts $\times$ 5 seeds.
All frequency ratios use a normalized threshold $\rho=0.25$; temporal ratios exclude the DC bin.

\begin{table}[t]
\centering
\small
\caption{Metric card for the qualitative noise-space analysis.}
\label{tab:metric_card}
\begin{tabular}{p{0.22\linewidth} p{0.74\linewidth}}
\toprule
Metric & Definition / notes \\
\midrule
RelDisp & $\|d\|_2/\|z\|_2$ on flattened tensors. \\
CosSim  & $\cos(z,z_g)$ on flattened tensors. \\
DirStab & Mean pairwise cosine similarity of unit displacements across $S=5$ seeds (Eq.~\ref{eq:dirstab_app}); computed per prompt, then averaged. \\
CV$_{\|d\|}$ & $\mathrm{Std}(\|d_s\|_2)/\mathrm{Mean}(\|d_s\|_2)$ over seeds; computed per prompt. \\
EVR1    & Top explained-variance ratio from PCA over $\{d_s\}_{s=1}^{5}$ (Eq.~\ref{eq:evr1_app}); computed per prompt. \\
sp\_hf($x$) & Spatial HF power ratio from $\mathrm{rFFT2}$ over $(H,W)$ averaged across $(c,t)$ (Eq.~\ref{eq:sphf_app}). \\
t\_hf($x$)  & Temporal HF power ratio from $\mathrm{rFFT}$ over $T$ averaged across $(c,h,w)$, excluding DC (Eq.~\ref{eq:thf_app}). \\
tDiffRel($x$) & Relative temporal-difference RMS: $\mathrm{RMS}(\Delta_t x)/\mathrm{RMS}(x)$ with $\Delta_t x(:,t)=x(:,t+1)-x(:,t)$. \\
\bottomrule
\end{tabular}
\end{table}

\noindent \textbf{Per-prompt aggregation over seeds.}
DirStab/CV/EVR1 are computed \emph{per prompt} over the 5 seeds, and then aggregated by mean/median across prompts.

\subsection{Directional stability and low-rank structure}
For a prompt, let $\{d_s\}_{s=1}^{S}$ be displacements across $S=5$ seeds, and define unit directions $u_s=d_s/\|d_s\|_2$.
Directional Stability is
\begin{equation}
\mathrm{DirStab} \triangleq \frac{2}{S(S-1)}\sum_{1\le i<j\le S} \langle u_i, u_j\rangle.
\label{eq:dirstab_app}
\end{equation}
To quantify low-rank structure, we run PCA on the set of flattened $\{d_s\}$ (after centering across seeds).
If $\{\lambda_i\}$ are PCA eigenvalues, the explained variance ratio of the top component is
\begin{equation}
\mathrm{EVR1} \triangleq \frac{\max_i \lambda_i}{\sum_i \lambda_i}.
\label{eq:evr1_app}
\end{equation}

\subsection{Spatial high-frequency ratio}
For $x\in\mathbb{R}^{C\times T\times H\times W}$, we compute a 2D real FFT over $(H,W)$ for each channel and time:
\begin{equation}
X(c,t,\omega_h,\omega_w) = \mathrm{rFFT2}\big(x(c,t,:,:)\big),\quad
P = |X|^2.
\label{eq:spfft_app}
\end{equation}
We average power over $(c,t)$ to obtain $\bar{P}(\omega_h,\omega_w)$ and define a normalized radial spatial frequency
\begin{equation}
r(\omega_h,\omega_w)=\sqrt{\left(\frac{\min(\omega_h, H-\omega_h)}{H/2}\right)^2+\left(\frac{\omega_w}{W/2}\right)^2 }.
\label{eq:radius_app}
\end{equation}
With threshold $\rho=0.25$, the spatial HF ratio is
\begin{equation}
\mathrm{sp\_hf}(x) \triangleq
\frac{\sum_{r\ge \rho}\bar{P}}{\sum \bar{P}}.
\label{eq:sphf_app}
\end{equation}

\subsection{Temporal high-frequency ratio and temporal-difference metrics}
For temporal HF ratio, we apply a 1D real FFT along time $T$ for each $(c,h,w)$:
\begin{equation}
Y(c,k,h,w) = \mathrm{rFFT}\big(x(c,:,h,w)\big),\quad
Q(k)=\mathbb{E}_{c,h,w}\big[|Y(c,k,h,w)|^2\big],
\label{eq:tfft_app}
\end{equation}
where $k=0,\ldots,K-1$ and $K=T/2+1$. Excluding the DC bin $k=0$, define normalized frequency $f_k = k/(K-1)$ and threshold $\rho_t=0.25$:
\begin{equation}
\mathrm{t\_hf}(x)\triangleq
\frac{\sum_{k>0,\, f_k\ge \rho_t} Q(k)}{\sum_{k>0} Q(k)}.
\label{eq:thf_app}
\end{equation}
To measure temporal discontinuity, we compute first-order differences
$\Delta_t x(:,t,:,:)=x(:,t+1,:,:)-x(:,t,:,:)$. We report
\begin{equation}
\mathrm{tDiffRMS}(x)\triangleq \sqrt{\mathbb{E}\big[(\Delta_t x)^2\big]},
\qquad
\mathrm{tDiffRel}(x)\triangleq \frac{\mathrm{tDiffRMS}(x)}{\sqrt{\mathbb{E}[x^2]}}.
\label{eq:tdiff_app}
\end{equation}

\end{document}